\documentclass[12pt]{article}
\usepackage{amsmath}
\usepackage{graphicx,authblk}
\usepackage{enumerate,color,rotating,booktabs,multirow,changepage,anyfontsize,algorithm2e,amsfonts}
\usepackage{url,subcaption} 

\newcommand{\blind}{0}

\DeclareMathOperator*{\argmin}{arg\,min}

\begin{document}

\def\spacingset#1{\renewcommand{\baselinestretch}%
{#1}\small\normalsize} \spacingset{1}


\if0\blind
{
  \title{\bf Multi-source domain adaptation for regression}
  \author{Yujie Wu\\
    \textit{Department of Biostatistics, Harvard University} \\
    Giovanni Parmigiani \\
    \textit{Department of Biostatistics, Harvard University,}\\
\textit{Department of Data Science, Dana-Farber Cancer Institute}\\
and Boyu Ren\\
\textit{Laboratory for Psychiatric Biostatistics, McLean Hospital}\\
\textit{Department of Psychiatry, Harvard Medical School}
}
\date{}
  \maketitle
} \fi

\bigskip
\begin{abstract}
Multi-source domain adaptation (DA) aims at leveraging information from more than one source domain to make predictions in a target domain, where different domains may have different data distributions. Most existing methods for multi-source DA focus on classification problems while there is only limited investigation in the regression settings. In this paper, we fill in this gap through a two-step procedure. First, we extend a flexible single-source DA algorithm for classification through outcome-coarsening to enable its application to regression problems. We then augment our single-source DA algorithm for regression with ensemble learning to achieve multi-source DA. We consider three learning paradigms in the ensemble algorithm, which combines linearly the target-adapted learners trained with each source domain: (i) a multi-source stacking algorithm to obtain the ensemble weights; (ii) a similarity-based weighting where the weights reflect the quality of DA of each target-adapted learner; and (iii) a combination of the stacking and similarity weights. We illustrate the performance of our algorithms with simulations and a data application where the goal is to predict High-density lipoprotein (HDL) cholesterol levels using gut microbiome. We observe a consistent improvement in prediction performance of our multi-source DA algorithm  over the routinely used methods in all these scenarios.
\end{abstract}

\newpage
\spacingset{1.9} 

\section{Introduction}
\label{intro}

In biomedical and clinical research, it has become increasingly common to have multiple data sets that measure the same outcomes $Y$ and covariates $X$ for synthesis analyses \cite{patil2018training, guan2019merging}. For example, Gene Expression Omnibus\cite{edgar2002gene} and ArrayExpress \cite{parkinson2010arrayexpress} contain gene expression data from over 70,000 studies while Alzheimer’s Disease Neuroimaging Initiative \cite{mueller2005alzheimer} publish multi-modal data, spanning from neuro-cognitive measures to imaging biomarkers and behavioral tests, for patients with Alzheimer's disease across multiple sites and times. Typically, the purpose of the synthesis effort is either to perform inference on the super-population, from which all these data sets are sampled (e.g., meta-analysis and domain generalization), or to leverage this rich data collection (i.e., source domains) to better inform the structure of a target domain, which can be represented by one of the data sets or an external data set containing covariates only. In this manuscript, we focus on the latter purpose in the context of predictions and deal with the case where we \textit{only} observe covariates in the target domain.

The main statistical challenge for our task is to handle the potential heterogeneity of data distributions across the various domains \cite{zhang2020impact} due to their distinct study populations, study designs and study-specific technological artifacts \cite{simon2003pitfalls, rhodes2004large, patil2015test, sinha2017assessment}. The presence of domain heterogeneity implies that prediction algorithms that work well for a source domain might have subpar performance for the target domain. This issue has been studied extensively in research of unsupervised domain adaptation (UDA) \cite{farahani2021brief} when only one source domain is available. The type of UDA that is directly relevant for our analysis is under the setting of Generalized Target Shift (GeTarS), where the joint distribution of the outcome and covariates $P(X,Y)$ is different across domains. A common strategy to solve GeTarS is to decompose the study heterogeneity into the differences of $P(Y)$ and $P(X|Y)$. The former difference, also known as Target Shift (TarS), is typically handled by reweighting the loss function of the source domain by the density ratio of $Y$. The estimation of the density ratio of $Y$ can be achieved even in absence of $Y$ in the target domain by matching on the marginal distributions of $X$ \cite{zhang2013domain, lipton2018detecting}. The difference of $P(X|Y)$, on the other hand, has been considered as a representation learning task in most of the existing literature. Specifically, the goal is to identify a transformation $\tilde X$ of the $X$ such that $P(\tilde X|Y)$ is the same across domains. Earlier work employs the simple location-scale (LS) transformation \cite{zhang2013domain, gong2016domain}, while the more flexible adversarial learning based approaches, where the transformation is induced through a deep neural network, have become more prevalent recently \cite{ganin2016domain, tachet2020domain, shui2021aggregating}.

The generalization of the UDA algorithms to the case of multiple source domains is non-trivial, mostly because of the burden of reconciliation of the complex heterogeneity structure across source domains. A string of methods for multi-source DA (MSDA) focus on training source specific learners and ensemble the prediction results, while the construction of ensemble weights often fails to acknowledge the difference in data distribution across domains \cite{peng2019moment, guo2018multi}. Moreover, most recent development of MSDA focuses primarily on classification problems, whereas the regression problems only receive limited attention \cite{zhao2020multi}. These two limitations combined poses a critical challenge in applying MSDA in biomedical and clinical research, where substantial diversity in study populations and study designs is ubiquitous, and the outcomes of interest, such as disease risks for prompt intervention and treatment, are often continuous \cite{meigs2008genotype, de2015current}.

We propose an ensemble learning framework for MSDA that allows for both continuous and discrete outcomes. Our framework is based on the cross-study stacking algorithm \cite{patil2018training, ramchandran2019tree, loewinger2022hierarchical}, which has been shown to have superior performance over merging when domain heterogeneity is high \cite{patil2018training, guan2019merging}. We first introduce an computationally efficient single-source DA algorithm for regression and then couple it with a novel cross-domain stacking approach to achieve MSDA in presence of study heterogeneity. The final prediction model combines a collection of predictors, trained by adapting each of the source domains to the target domain, with weights selected by maximizing a utility function that promotes high potential of generalizability across domains. The paper is organized as follows. In Section \ref{methods}, we introduce the approach to single-source DA for regression 
and its extension to the multi-source setting; in Section \ref{simulation}, we evaluate the performance of our methods through extensive simulation studies; in Section \ref{real_data}, we apply the methods to predict cholesterol levels based on metagenomic sequencing data from three different real world studies and Section \ref{discussion} concludes the paper.


\section{Methods}
\label{methods}
\subsection{Problem setups}
Let $\mathcal{X}$ and $\mathcal{Y}$ denote the input and output spaces. We consider the regression problem where $\mathcal{Y}\subset \mathbb R$ is a continuous domain. We use $\mathcal{S}_1, \mathcal{S}_2,\ldots, \mathcal{S}_K$ to denote the density of $(X,Y)$ in source domains and $\mathcal{T}$ for the target domain. Let the density ratio of $Y$ between the target domain $\mathcal T$ and a source domain $k$ be $\beta_k(y) = \mathcal T(y)/\mathcal S_k(y)$. The sample sizes for these domains are $n_1,n_2,\ldots, n_K$ and $n_{\mathcal{T}}$, respectively. Following \cite{zhang2013domain}, we assume the marginal distributions $\mathcal{S}_k(y)\ne\mathcal{T}(y)$ and the conditional distributions $\mathcal{S}_k(x \mid y)\ne\mathcal{T}(x\mid y)$ for $k=1,\ldots,K$. Our algorithm consists of two main steps: 1) learn the source-specific predictors $G_y^k: \mathcal{X}\times\mathcal{Y}\to \mathbb R$ that minimize the risk with respect to the target domain distribution: $\mathbb{E}_{({x},y)\sim \mathcal{T}}\left[\ell(G_y^k({x}), y)\right]$, where $\ell$ is some prediction loss functions; 2) combine the source-specific predictors $G_y^k$ through stacking to form the final predicting rule $G_y=\sum_k w_kG_y^k$ that leverages information from all source domains, where $w_k$ is the ensemble weight for the $k$-th source domain.

\subsection{Single-source DA for source-specific predictors $G_y^k$}
\label{single_da}

In this section, we focus on the algorithm to derive $G_y^k$ for a specific $k$ through DA. We adopt the approach in Zhang et al. \cite{zhang2013domain} and deal with the shift of $\mathcal S_k(y)$ and $\mathcal S_k(x|y)$ between a source domain and the target domain separately. We account for the target shift through importance sampling and for the shift in $\mathcal S_k(x|y)$, an adversarial learning approach is used. We define the expected loss function for $G_y^k$ that explicitly incorporates both strategies as follows.
$$
\mathbb{E}_{({x},y)\sim \mathcal{S}_k}\left[\beta_k(y)\ell(G_y^k(G_f^k({x})), y)\right].
$$
Here $\beta_k(y)$ is the importance weight of observations in source domain $k$ with respect to the target domain and $G_f^k$ is the feature representation such that $\mathcal S_k(G_f^k(x)|y)=\mathcal T(G_f^k(x)|y)$. We learn $\beta_k(y)$ and $G_f^k(\cdot)$ through an iterative process: $\beta_k$ is estimated using the $G_f^k(\cdot)$ from the previous iteration under the assumption that this $G_f^k(\cdot)$ has already corrected for the shift in $\mathcal S_k(x|y)$ and $G_f^k(\cdot)$ is derived by fixing $\beta_k$ at its current estimated values. We now discuss these two components of the estimation procedure in details.

\subsubsection{Estimating $\beta_k(y)$ for regression}

We discuss the approach for estimating importance weights $\beta_k(y)$ when $y$ is continuous. Since we only consider a single source domain, we omit the subscript $k$. The approach is an extension of the Black Box Shift Estimation (BBSE) for classification. The BBSE relies on three working assumptions: 1) the conditional distributions of the features given the outcome are the same: $\mathcal{S}({x}|y)=\mathcal{T}({x}|y)$, $\forall {x}$ and $y$; 2) $\forall y\in\mathcal{Y}$ with $\mathcal{T}(y)>0$, $\mathcal{S}(y)>0$; 3) the expected confusion matrix of the black box predictor discussed below is invertible. Note that, Assumption 1 does not hold in our setting for the original features $X$, but it does for the transformed features $G_y^k(x)$ after sufficient rounds of iterations. This enables us to apply BBSE directly by replacing the original features with the transformed ones \cite{garg2020unified}. 

In BBSE for discrete outcomes with $L$ categories, a black box predictor ($f(x)$) such as random forest or neural network will first be trained in the source domain and the importance weights are estimated by solving the following linear systems of equations:
\begin{equation}
	{\mu}_{\widehat{y}}={C}_{\widehat{y},y}{\beta}(y)
	\label{BBSE}
\end{equation}
where ${\mu}_{\widehat{y}}$ is a $L$-dimensional column vector with the $l$-th element $[{\mu}_{\widehat{y}}]_l=\mathcal{T}(f(x)=l)$, representing the prediction mass of the $l$-th category from the black box predictor in the target domain, while ${C}_{\widehat{y},y}$ is the $L\times L$ confusion matrix of the black box predictor in the source domain, with the $(i,j)$-th element $[{C}_{\widehat{y},y}]_{(i,j)}=\mathcal{S}_k(f(x)=i, y=j)$ and ${\beta}(y)$ is a $L$-dimensional column vector for the importance weights over the $L$ categories. We refer the readers to \cite{lipton2018detecting} for technical details. However, Equation (\ref{BBSE}) fails to ensure the weights are non-negative and sums up to 1 \cite{gretton2009covariate}, we therefore propose the following constrained optimization:

\begin{equation}
	\begin{split}
		\arg\min_{{\beta}(y)}&||{\mu}_{\widehat{y}}-{{C}}_{\widehat{y},y}{\beta}(y)||^2_2\\
		\text{s.t.} &\quad \beta(y)\ge0\text{ and }\Big|{1}^T{\beta}(y)-1\Big|<\epsilon,\\
	\end{split}
	\label{BBSE_opt}
\end{equation}
where $\epsilon$ is some small value. Optimizing Equation (\ref{BBSE_opt}) is a standard constrained quadratic programming problem that can be easily solved using existing software.

When $y$ is continuous, ${\beta}(y)$ becomes a continuous function of $y$, and we choose to model ${\beta}(y)$ using restricted cubic splines. As a result, the importance weight for the $i$-th observation is represented as $\beta(y_i;{\alpha})=\sum_{m=1}^M\alpha_mq_m(y_i)$, where $q_m(\cdot)$ is the basis function for restricted cubic splines, and $M$ is determined by the number of knots. In our experiments, we use restricted cubic splines with 12 knots, and the location of knots are based on quantiles of $Y$ in the source domain [\cite{harrell2001regression}]. Note that for the $k$-th source domain, ${\mu}_{\widehat{y}}$ and ${C}_{\widehat{y},y}$ are a $n_k$-dimensional column vector and a $n_k\times n_k$-dimensional confusion matrix, and the coefficients $\alpha_m$ corresponding to the basis functions could be estimated by minimizing the objective function in Equation (\ref{BBSE_opt}). However, this approach is computationally expensive even under moderate sample sizes. We instead choose to categorize the outcomes into $L$ levels for scalability based on thresholds such as quantiles of $Y$'s in the source domain, and ${\beta}(y)$, ${\mu}_{\widehat{y}}$ and ${C}_{\widehat{y},y}$ become $L$-dimensional column vectors and a $L\times L$ matrix regardless of the sample size. The $l$-th element in ${\beta}(y)$ can be interpreted as the mean of the actual $\beta(y)$ across the range of $y$ associated with the $l$-th category. In other words, $[{\beta}(y;{\alpha})]_l\approx \frac{1}{\#:y_i\in\text{Category  }l}\sum_{i: y_i\in\text{Category } l}\beta(y_i;{\alpha})$. Estimation of the coefficients ${\alpha}$ can be similarly carried out as in the case of classification by replacing $\beta(y)$ in (\ref{BBSE_opt}) with $\beta(y,\alpha)$. We summarize the extended BBSE algorithm for regression in Algorithm \ref{algo:BBSE}.

\RestyleAlgo{ruled}
\begin{algorithm}
	\caption{Extended BBSE algorithm}\label{algo_BBSE}
	\KwData{${\mathcal{S}}_k$ ($k$-th source domain), ${\mathcal{T}}$ (target domain)}
	\KwResult{ $\widehat{\beta}(y_i)$ for $i=1,\ldots,n_k$}
	\textbf{Initialization}: Select the number of categories $L$. We recommend $L$ such that $n_k/L^2\ge 5$. Create the corresponding categorical outcomes based on pre-specified cutoff values of user's choice or quantiles of $Y$ in the source domain.
	
	\textbf{Step 1. }Split the source domain, and use the first fold to train the black box predictor;
	
	\textbf{Step 2.} Compute the estimated confusion matrix $\widehat{{C}}_{\widehat{y},y}$ in the second fold of the source domain and $\widehat{{\mu}}_{\widehat{y}}$ in the target domain;
	
	\textbf{Step 3.} Optimize Equation (\ref{BBSE_opt}) and get the estimated weights $\widehat{\beta}(y_i)=\sum_{m=1}^M\widehat{\alpha}_mq_m(y_i)$. 
	
	\label{algo:BBSE}
	
\end{algorithm}

\subsubsection{Adversarial learning to identify $G_y^k(\cdot)$}

To deal with the difference in the conditional distributions of $X$ given $Y$ between the $k$-th source domain and target domain, a source-specific transformation function $G_f^k(\cdot): \mathcal{X}\to\mathcal{Z}$ will be learned and applied to features in both the source and target domains, such that $\mathcal{S}_k(G_f^k({x})|Y)=\mathcal{T}(G_f^k({x})|Y)$, and the final source-specific predicting model $G_y^k()$ takes the transformed features as input. The feature transformation function $G_f^k()$ can be learned following an adversarial learning framework \cite{shui2021aggregating} through minimizing the weighted implicit conditional Wasserstein distance. In practice, $G_y^k()$ and $G_f^k()$ are jointly trained by optimizing the following objective function:

\begin{equation}
	\argmin\limits_{G_y^k, G_f^k}\max_{d_k}	\underbrace{\frac{1}{n_k}\sum_{i=1}^{n_k}\beta(y_i)[y_i-G_y^k(G_f^k(x))]^2}_{\mathcal{L}_R}+\underbrace{\lambda\left[\frac{1}{n_k}\sum_{i=1}^{n_k}{\beta}(y_i)d_k(G_f^k(x_i))-\frac{1}{n_{\mathcal{T}}}\sum_{j=1}^{n_{\mathcal{T}}}d_k(G_f^k(x_j))\right]}_{\mathcal{L}_{DA}}.
	\label{rl}
\end{equation}
The objective function can be decomposed into two components. The first component, $\mathcal{L}_R$, is the reweighted regression loss in the $k$-th source domain, which is the empirical estimate of the risk $\mathbb{E}_{({x},y)\sim \mathcal{S}_k}\left[\beta_k(y)\ell(G_y^k(G_f^k({x})), y)\right]$, and the second component, $\mathcal{L}_{DA}$, quantifies the implicit conditional Wasserstein-1 distance of the transformed features between the source and target domains, in which $d_k()$ is the 1-Lipschitz domain discriminator for the $k$-th source domain, and we refer the readers to \cite{ganin2016domain,shui2021aggregating} for technical details.

Note that, Equation (\ref{rl}) involves the importance weights $\beta(y)$ that should be estimated following our proposed extended BBSE algorithm described in last section, while the extended BBSE algorithm also relies on the transformed features. Therefore, in practice, for general DA where both $\mathcal{S}_k(y)\ne\mathcal{T}(y)$ and  $\mathcal{S}_k({x}| y)\ne\mathcal{T}({x}| y)$, our single-source DA algorithm will iterate between optimizing Equation (\ref{BBSE_opt}) and Equation (\ref{rl}) to gradually align the conditional distributions as well as estimating the importance weights. We describe our single-source DA algorithm for regression in Algorithm \ref{algo:SSDA}.

\RestyleAlgo{ruled}

\begin{algorithm}
	\caption{Single-source DA for regression}\label{alg:core}
	\KwData{${\mathcal{S}}_k$ ($k$-th source domain), ${\mathcal{T}}$ (target domain)}
	\KwResult{ $\widehat{G}_f^k(\cdot), \widehat{G}_y^k(\cdot)$ }
	\While{convergence criteria not met}{
		\textbf{Step 1.} Run the extended BBSE algorithm using the transformed features $\widehat{G}_f^{k}({x})$ from the last iteration to obtain updated importance weights $\widehat{\beta}(y)$;
		
		\textbf{Step 2.} Optimize Equation (\ref{rl}) using updated weights $\widehat{\beta}(y_i)$ from Step 1, and obtain the updated feature transformation function $\widehat{G}_f^{k}(\cdot)$ and the final predicting model $\widehat{G}_y^{k}(\cdot)$
	}
	
	\label{algo:SSDA}

\end{algorithm}

\subsection{Multi-source domain adaptation}
We extend our single-source DA to the scenario where multiple source domains are available through stacking \cite{patil2018training}. In the first phase of this approach, we apply the single-source DA algorithm in Algorithm \ref{algo:SSDA} for each source and target domain pair $(\mathcal{S}_k, \mathcal{T}), k=1,\ldots, K$. We then linearly combine the source-specific predictions function $G_y^k\circ G_f^k$ to obtain the final prediction model $G_y^\mathcal{T}$ for the target domain:
$$
G_y^\mathcal{T} = \sum_k w_k G_y^k\circ G_f^k.
$$

This ensemble approach is effective in reducing prediction variance and yielding more accurate predictions than each of the ensemble members\cite{sagi2018ensemble, hastie2009elements}. In the following sections, we introduce three methods to obtain the combination weights $w = (w_1,\ldots,w_K)$.

\subsubsection{Multi-source stacking weights}
\label{stacking}

We obtain the weights through the stacking regression \cite{patil2018training}, where the source-specific predictors will be applied to make predictions on all source domains, and the weights are obtained by regressing the true outcomes from the source domains against the predicted values from the $K$ source-specific predictors. When predicting outcomes for source $k$ using data of source $k'$, we replace $G_y^{k'}\circ G_f^{k'}$ with $G^{\mathcal S_{k'},\mathcal S_{k}}_y\circ G^{\mathcal S_{k'},\mathcal S_{k}}_f$, which is the adapted prediction function for domain $\mathcal{S}_k$ based on data in domain $\mathcal{S}_{k'}$. In other words, we learn $G^{\mathcal S_{k'},\mathcal S_{k}}_y$ and $G^{\mathcal S_{k'},\mathcal S_{k}}_f$ by invoking Algorithm \ref{algo:SSDA} with $S_{k'}$ being the source domain and $S_{k}$ the target domain. We can write the design matrix of the stacking regression problem as follows.

$$
\hat Y = 
\begin{bmatrix}
	{G}_y^{\mathcal{S}_1}({X}_{\mathcal{S}_1})&{G}_y^{\mathcal{S}_2,\mathcal{S}_1}({G}_f^{\mathcal{S}_2,\mathcal{S}_1}({X}_{\mathcal{S}_1}))&\ldots&{G}_y^{\mathcal{S}_K,\mathcal{S}_1}({G}_f^{\mathcal{S}_K,\mathcal{S}_1}({X}_{\mathcal{S}_1}))\\
	{G}_y^{\mathcal{S}_1,\mathcal{S}_2}({G}_f^{\mathcal{S}_1,\mathcal{S}_2}({X}_{\mathcal{S}_2}))&{G}_y^{\mathcal{S}_2}({X}_{\mathcal{S}_2})&\ldots&{G}_y^{\mathcal{S}_K,\mathcal{S}_2}({G}_f^{\mathcal{S}_K,\mathcal{S}_2}({X}_{\mathcal{S}_2}))\\
	\vdots&{\color{black}{\vdots}}&&\vdots\\
	{G}_y^{\mathcal{S}_1,\mathcal{S}_K}({G}_f^{\mathcal{S}_1,\mathcal{S}_K}({X}_{\mathcal{S}_K}))&{G}_y^{\mathcal{S}_2,\mathcal{S}_K}({G}_f^{\mathcal{S}_2,\mathcal{S}_K}({X}_{\mathcal{S}_K}))&\ldots&{G}_y^{\mathcal{S}_K}({X}_{\mathcal{S}_K})\\
\end{bmatrix}
$$
We obtain the combination weights $w_{stack}=(w_{stack,1},\ldots, w_{stack,K})$ with the following optimization:
\begin{equation}
	\hat w_{stack} = \argmin\limits_{w\in \mathbb S_{K-1}} \|Y - \hat Y w\|_2^2,
	\label{stacking_opt}
\end{equation} 
where $Y = (Y_1,\ldots, Y_K)$ is the concatenated outcome vector across all $K$ source domains and $\mathbb S_{K-1}$ is the standard simplex. The resulting weight $w_{stack, k}$ is large when the source domain $k$ generates high quality predictions for all other source domains using our DA algorithm. The weights $w_{stack}$ thus reflects the general adaptation and generalizability of the $k$-th source domain to other domains. In the final prediction model $G^\mathcal T$, we prioritize the adapted prediction models $G_y^k\circ G_f^k$ of source domains with high generalizability, which are also more likely to achieve successful DA to the target domain. Algorithm \ref{stack_algo} summarizes our proposed multi-source stacking DA method.

\RestyleAlgo{ruled}

\begin{algorithm}[h!]
	\caption{Multi-source stacking domain adaptation}\label{stack_algo}
	\KwData{${\mathcal{S}}_1,\ldots, {\mathcal{S}}_K$ (source domains), ${\mathcal{T}}$ (target domain)}
	\For{$i = 1, 2, \ldots, K$}{
		\For{$j = 1, 2, \ldots, K$}{
			\If{$i\ne j$ }{
				$\mathcal{S}_i$ as the `source' domain and $\mathcal{S}_j$ as the `target' domain, and run the single-source DA algorithm, obtaining the predictions: $\widehat{G}_y^{\mathcal{S}_i,\mathcal{S}_j}(\widehat{G}_f^{\mathcal{S}_i,\mathcal{S}_j}({X}_{\mathcal{S}_j}))$.
				
				Fill in the $(i,j)$-th block of the stacking design matrix $\widehat{{Y}}$.
			}\Else{
				Train a traditional machine learning algorithm of user's choice on $\mathcal{S}_i$ and predict in $\mathcal{S}_i$.
				
				Fill in the $(i,i)$-th block of the stacking design matrix.
			}
	}}
	
	Optimize Equation (\ref{stacking_opt}) to obtain the combination weights: $\widehat{w}_{stack, 1}, \widehat{w}_{stack, 2},\ldots, \widehat{w}_{stack, K}$;
	
	\For{$i = 1, 2, \ldots, K$}{
		Perform the single-source domain adaptation between $\mathcal{S}_i$ and $\mathcal{T}$, obtaining the predictions in the target domain: $\widehat{G}_y^{k}(\widehat{G}_f^{k}({X}_{\mathcal{T}}))$.
	}
	
	\textbf{Output:} The final predictions are the following ensemble: $\sum_{k=1}^{K}\widehat{w}_{stack, k}\widehat{G}_y^{k}(\widehat{G}_f^{k}({X}_{\mathcal{T}}))$

\end{algorithm}

\subsubsection{Similarity-based ensemble weights}
\label{similarity}
Apart from using the stacking strategy in Section \ref{stacking} to obtain the ensemble weights, a simpler ensembling strategy is to directly consider how each source domain is adapted to the target domain, and the source domain with better adaptation should receive larger weight. A good candidate that reflects the quality of DA to the target domain is the weighted regression loss $\mathcal{L}_R$ from the objective function in Equation (\ref{rl}), where a smaller value indicates smaller target domain risk of the corresponding predictor. Let $\widehat{J}_{\mathcal{S}_k}$ denote the weighted regression loss for the $k$-th source domain obtained from the single-source DA algorithm, the following similarity-based weights can therefore be used for ensembling:
\begin{equation*}
	\widehat{w}_{sim,k}=\frac{1/\widehat{J}_k}{\sum_{l=1}^{K}1/\widehat{J}_l}, k=1,\ldots, K.
\end{equation*}

\subsubsection{Combining stacking weights with similarity weights}

It is natural to combine the stacking weights in Section \ref{stacking} with the similarity weights in Section \ref{similarity} to achieve robustness against extreme scnearios. For example, if a majority of the $K$ source domains are identical but different from the target domain, stacking will erroneously assign large weights to them and disregard the remaining source domains, even if some of these left-out source domains are indeed identical to the target domain. In this scenario, the similarity weights is preferred. On the other hand, if all source domains are adapted to the target domain equally well, which can happen if the source and target domains are sampled from a common hyper-distribution, the stacking weights might be preferred over the similarity weights. The two weighting strategy can be combined by solving the following optimization when obtaining the stacking weights:
\begin{equation}
	\argmin_{w\in\mathbb S_{K-1}}({\color{black}{1-\gamma}})\|{\widehat{Y}}w -{Y}\|^2 + {\color{black}{\gamma}}\|w - w_{sim}\|^2,
	\label{combine}
\end{equation}
where $\gamma$ is a parameter controlling the relative importance of stacking weights and similarity weights. $\gamma$ cannot be selected through cross-validation since the target domain has different data distributions from the source domains. One strategy to overcome this issue is to inspect the difference in the similarity weights $\widehat{{W}}_{sim}$. Consider the gap of the importance weights $\max(\widehat{{W}}_{sim})-\min(\widehat{{W}}_{sim})$. A large gap indicates that among the $K$ source domains, some source domains are adapted to the target domain much better than some other source domains and thus more importance should be attached to the similarity weights; while a small difference indicates that all source domains have similar adaptation performance and thus the stacking method is preferred to obtain the ensemble weights. We therefore propose to use $\gamma=\max(\widehat{{W}}_{sim})-\min(\widehat{{W}}_{sim})$ as the hyper-parameter value for Equation (\ref{combine}).

\section{Simulation}
\label{simulation}
\subsection{Estimation of importance weights for regression}
In this section, we explore the performance of our extended BBSE algorithm for estimating the importance weights when the outcome is continuous. We generate one source domain and one target domain where the outcomes in the source domain are generated from $N(0, 1)$, with the outcomes in the target domain following $N(0.5, 0.5)$, and $X=Y+N(0, 0.5)$ for both the source and target domains. The sample sizes for both domains are 600, and 100 simulation replicates are performed. We set the black box predictor to be a multinomial logistic regression and we discretize outcomes into 4 categories and the thresholds are determined by quantiles of $Y$ in the source domain with equal space. To approximate the functional form of the importance weights $\beta(y)$, we use a restricted cubic spline with 12 knots. The baseline method is set to be an ordinary least squares (OLS) by regressing $Y$ against $X$ without any weights. We then fit a weighted least squares (WLS) using weights learned from our extended BBSE algorithm to deal with the shift in marginal distributions of $Y$.


\begin{figure}
	\centering
	\begin{subfigure}[b]{0.45\textwidth}
		\centering
		\includegraphics[width=\textwidth]{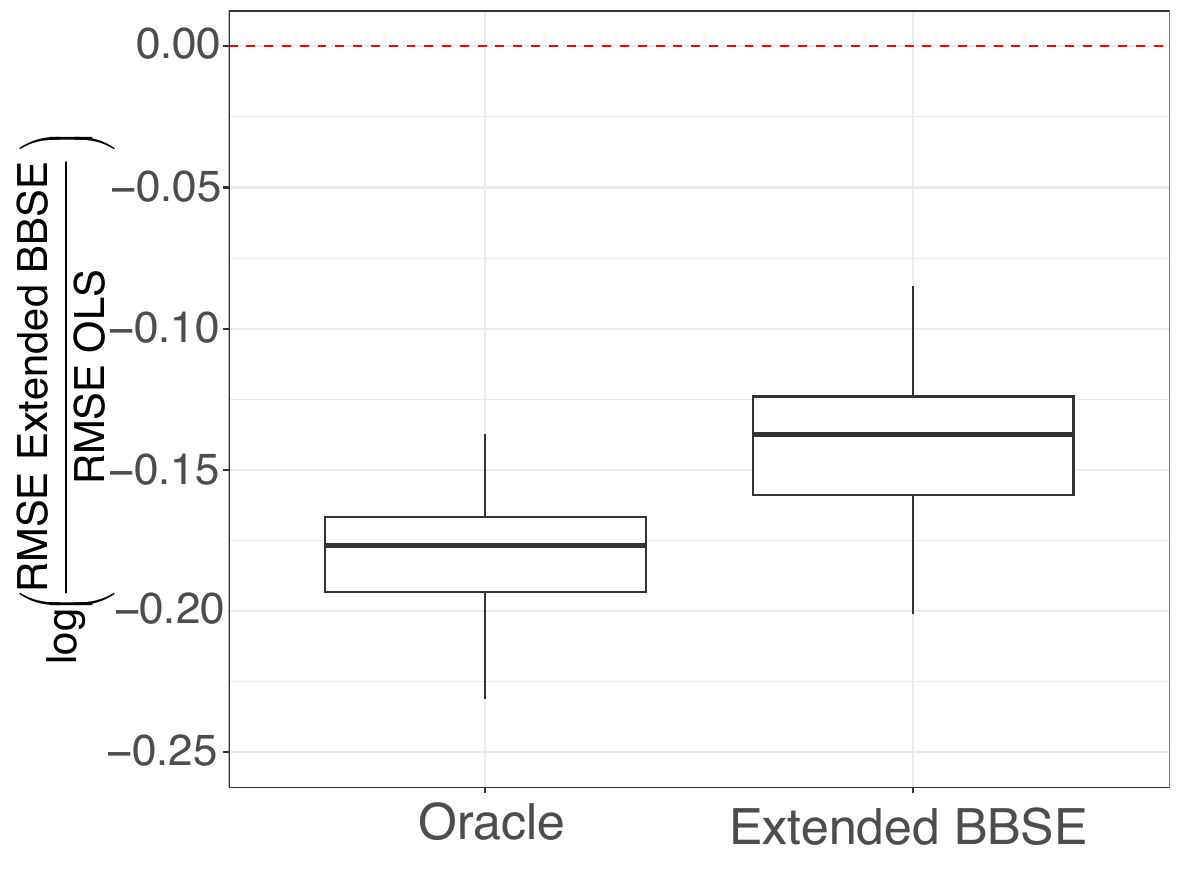}
		\caption{Log RMSE ratio for the extended BBSE algorithm}
		\label{1bbselinear}
	\end{subfigure}
	\hspace{1em}
	\begin{subfigure}[b]{0.45\textwidth}
		\centering
		\includegraphics[width=\textwidth]{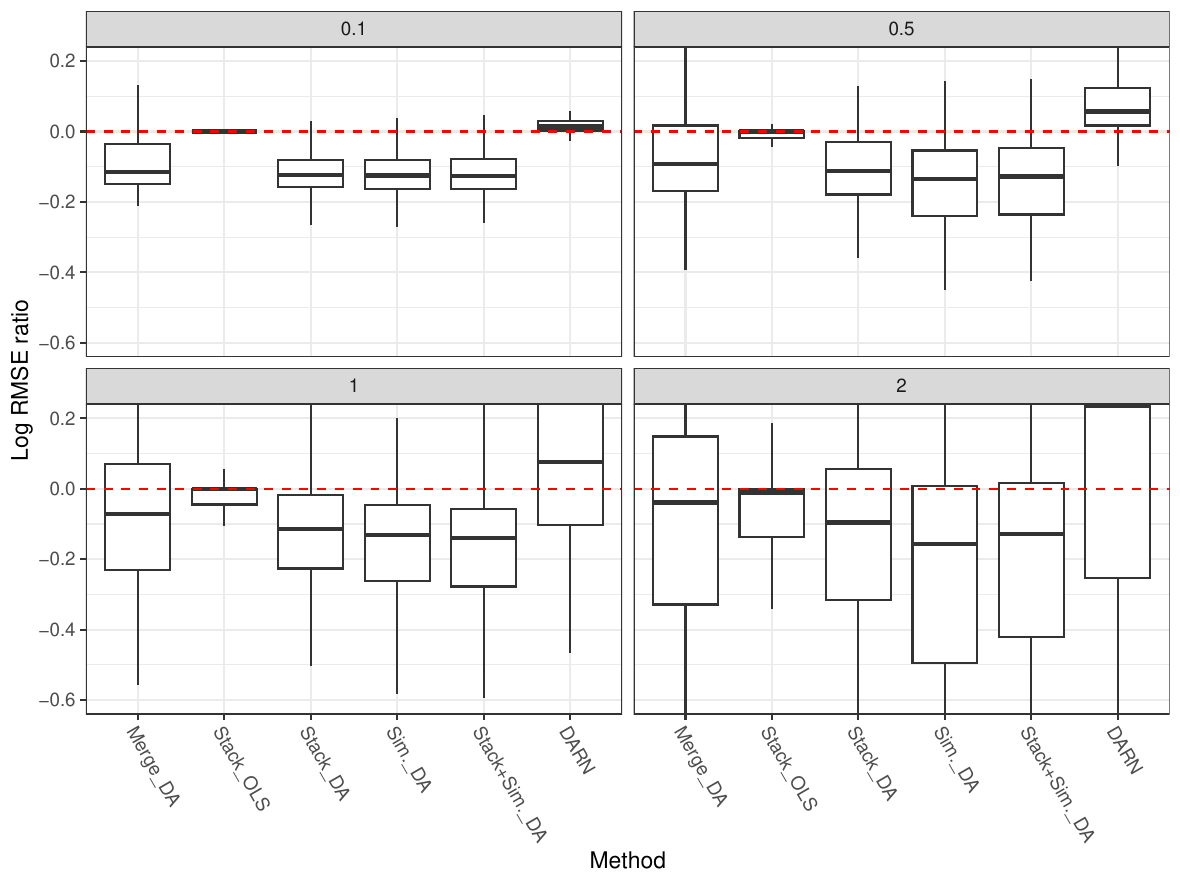}
		\caption{Log RMSE ratio for multy-source domain adaptation}
		\label{2multitanh}
	\end{subfigure}
	\caption{Simulation results}
\end{figure}

Figure \ref{1bbselinear} shows the boxplot of the log root mean squared error (RMSE) ratio of prediction in the target domain between our extended BBSE method and the baseline method without DA. A negative value corresponds to smaller prediction RMSE from our method. In addition, we plot the Oracle log RMSE ratio where we use the true weights when fitting the WLS. From the plot, we observe that our extended BBSE algorithm yields significant improvement of prediction accuracy in the target domain as compared with the baseline model without DA.

In addition, we consider two scenarios with more complicated data generation mechanism: (1) The outcomes are generated following the same way mentioned above, while the covariate is generated as: $X=\sin(Y)+N(0,0.5)$ for both the source and target domains. (2) The outcome is generated from a mixture distribution in the target domain, where $Y\sim 0.2N(0.2, 0.5) + 0.8N(1,1)$, while the outcome from the source domain is generated from $N(0,2)$. The covariate is generated as: $X=Y+3\tanh(Y)+N(0,1.5)$ for both the source and target domains. For both scenarios, the covariate-outcome relationship is not linear, and therefore we add a quadratic term of the covariate into the predicting model. The results can be found in supplementary material, where we observe similar behavior that our extended BBSE algorithm consistently yield smaller prediction RMSE than the baseline model without DA.


\subsection{Multi-source prediction}


In this section, we explore the performance of our proposed multi-source DA methods. We generate three source domains and one target domain. The outcomes in the source domains are generated from a hierarchical model: $Y_k\sim N(\mu_k, 1), k=1,\ldots,K$ with $\mu_k\sim N(0, \sigma^2)$, where $\sigma$ controls the study heterogeneity, and we set $\sigma=0.1, 0.5, 1$ and $2$ for varying degrees of study heterogeneity. In the target domain, we fix the outcome distribution as $Y\sim N(0.5, 0.5)$ to ensure the assumption for target shift is satisfied. The covariate for all source and target domains is generated from a hierarchical model, and for the $k$-th domain, we have: $X_k=\beta_{1,k}Y_k+\beta_{2,k}\tanh(Y_k)+N(0,1)$, with $\beta_{1,k}\sim N(1,\sigma^2)$ and $\beta_{2,k}\sim N(2,\sigma^2)$ for $k=1,\ldots, K, K+1$, where we use $k=K+1$ to denote the target domain. {\color{black}{When training the stacking weights for our proposed multi-source stacking DA algorithm, we further add the mean of all source domains as an additional single learner to calibrate the predictions. The added mean can be regarded as applying the DA algorithm on the merged data set, while the feature transformation function will map all the covariates to a constant.}}  In addition to using our proposed methods for multi-source DA: stacking (Stack\_DA), similarity weighting (Sim\_DA) and the combination of stacking with similarity weights (Stack+Sim\_DA), we consider merging all the source domains together, treating it as one source domain and run our proposed single-source DA algorithm on the merged data set (Merge\_DA), the usual stacking method where we fit separate regression models on each source domain without DA and ensemble the prediction results using the stacking technique (Stack\_OLS) \cite{patil2018training} {\color{black}{and one of the state-of-art methods Domain AggRegation Network (DARN) for multi-source domain adaptation that can deal with regression problems} \cite{wen2020domain}}.


Figure \ref{2multitanh} shows the boxplots of the log RMSE ratio of prediction in the target domain for different methods under varying degrees of study heterogeneity, and the baseline method for comparison is the merging approach, where we merge all source domains together and train a regression model without any DA. We observe that our proposed three multi-source DA methods consistently have smaller prediction RMSE than the baseline method, even when study heterogeneity is large ($\sigma=2$). The Stack\_DA method has slightly higher prediction RMSE than the Sim\_DA method but the variance is smaller especially under large study heterogeneity. The combination of stacking and the similarity weights, as a consequence, has prediction RMSE between the Stack\_DA and Sim\_DA method, and the prediction variance is also smaller than the Sim\_DA method. Moreover, the Merge\_DA method also shows a decrease in the prediction RMSE when study heterogeneity is small, while the decrease becomes negligible as study heterogeneity increases. We hypothesize that when study heterogeneity is large, merging all source domains together may lead to multimodal distributions on the merged dataset, rendering the single-source DA algorithm to struggle. Further, the Stack\_OLS method has nearly identical performance as the baseline method under small study heterogeneity settings, and when study heterogeneity is large, it has a tendency for yielding smaller prediction RMSE, although the median RMSE is still close to the baseline method. {\color{black}{The DARN method, however, performs poorly across different scenarios, and when study heterogeneity is large ($\sigma \ge 0.5$), the prediction performance is even worse than the baseline merging approach.}}





\section{Real data application}
\label{real_data}

Numerous predictive models based on  metagenomic sequencing data have been developed thanks to the ongoing research into associations between the gut microbiome and health-related outcomes \cite{fu2015gut, gupta2020predictive}. Prediction of health status based on metagenomic data can facilitate precision medicine on prompt intervention and treatment. In this paper, we apply our methods to predict cholesterol levels, a strong risk predictor for cardiovascular disease, based on metagenomic sequencing data, where previous studies have found that microbiome is associated with blood cholesterol levels \cite{le2019intestinal, kenny2020cholesterol}. We use datasets from the  \texttt{curatedMetagenomicData} R package \cite{pasolli2017accessible} containing multiple curated, uniformly processed whole-metagenome sequencing studies, among which three studies have cholesterol measurements available: (1) A study conducted among European women with normal, impaired or diabetic glucose control \cite{karlsson2013gut}, and we refer to this study as Karlsson; (2) A study conducted among families with history of type-I diabetes \cite{heintz2016integrated}, and we refer to it as Heintz-Buschart, and (3) A study conducted among Chinese type-II diabetes patients with non-diabetic controls \cite{qin2012metagenome}, and we name this study as Qin. In our analysis, we restrict the prediction among female participants and the sample sizes are 145, 32 and 151, respectively.

\begin{figure}
	\centering
	\begin{subfigure}[b]{0.45\textwidth}
		\centering
		\includegraphics[width=\textwidth]{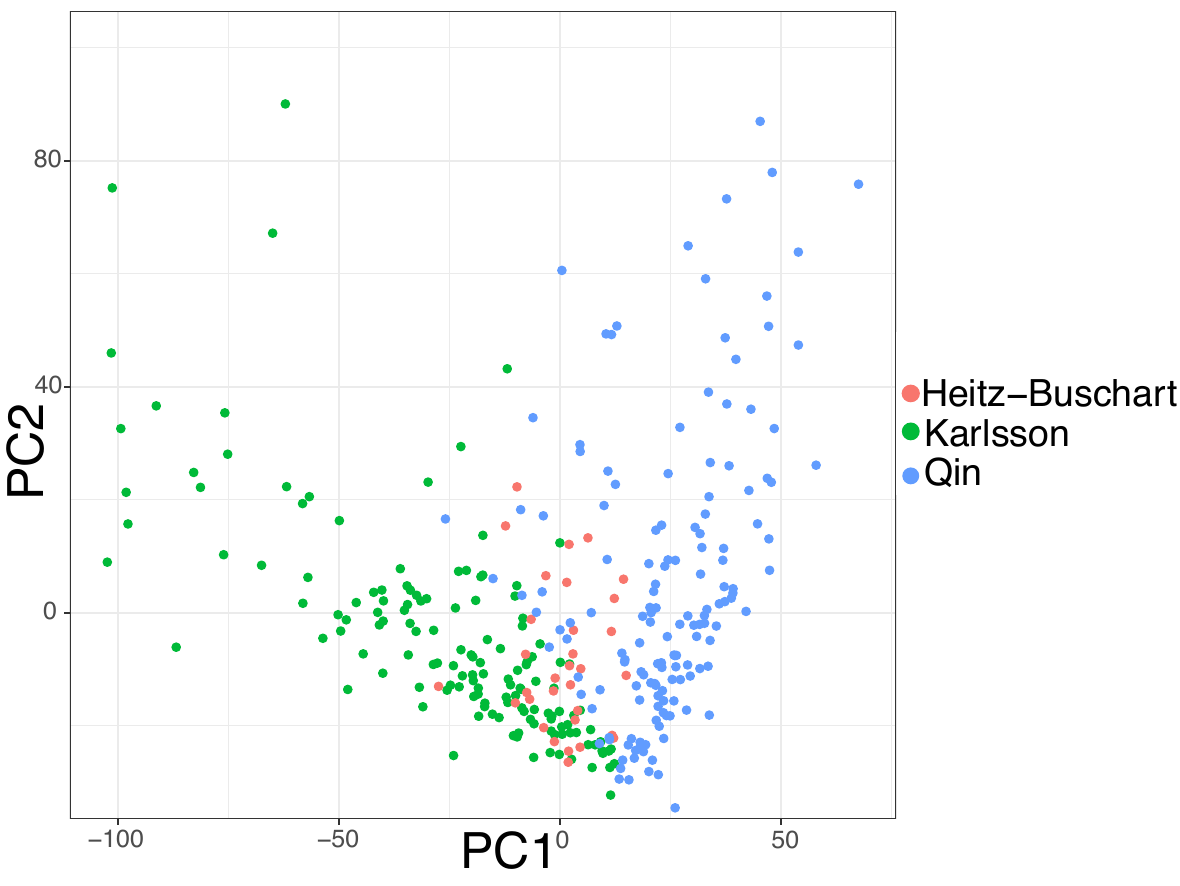}
		\caption{First two PCs of the marker abundance}
		\label{3pc}
	\end{subfigure}
	\hspace{1em}
	\begin{subfigure}[b]{0.45\textwidth}
		\centering
		\includegraphics[width=\textwidth]{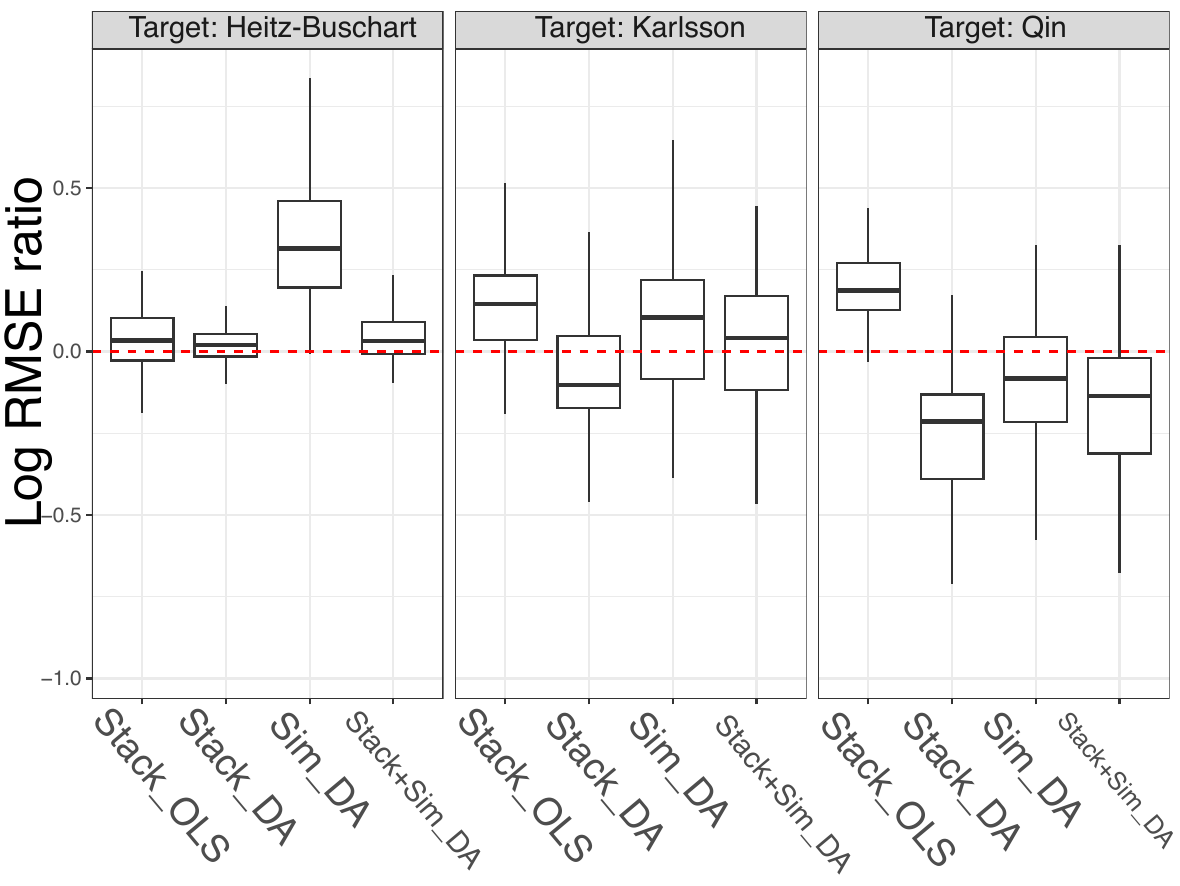}
		\caption{Bootstrap log RMSE ratio in the target domain}
		\label{3real}
	\end{subfigure}
	\caption{Real data application results}
\end{figure}

We use the gene marker abundance levels for prediction. Due to high dimensionality of the gene markers, we merge the gene markers across three studies and perform a principal component analysis (PCA) on the marker abundances, and the top 10 principal components (PC) with largest variances are used to build the prediction model for cholesterol levels. The first two PCs are plotted and annotated by study labels in Figure \ref{3pc}. We observe that the Qin and Karlsson studies are well separately by the first two PCs, suggesting the existence of study heterogeneity between the two studies, while the Heitz-Buschart study lies between the Qin and Karlsson studies.


To evaluate the performance of our proposed methods, we took two studies as the source domains and the remaining study as the target domain. Table \ref{real_rmse} reports the prediction RMSE. We observe that when the Qin or Karlsson study is used as the target domain, our proposed multi-source stacking DA algorithm yields the smallest prediction RMSE, while when the Heitz-Buschart study is used as the target domain, all the methods have worse prediction performance than the baseline merging method where we merge the source domains together and train a prediction model without DA. From the PC plot in Figure \ref{3pc}, we hypothesize that since the Heitz-Buschart study lies between the Karlsson and Qin study, merging these two source domains together may already contain all the useful information for prediction and additional DA may not be helpful but instead will introduce additional noises, rendering slightly larger prediction RMSE. However, contrary to the simulation results, using similarity weighting is less stable in real data application and has worse prediction performance than the baseline merging method when Karlsson or Heitz-Buschart study is used as the target domain. 


\begin{table}[h!]
	\caption{Prediction RMSE in the target domain}
	\begin{tabular}{cccccc}
		\hline
		Target domain		& Merging OLS & Stack OLS & Stack DA &  Similarity DA& Stack + Similarity DA  \\
		\hline
		Qin&52.64  & 51.35 & \textbf{37.02} &42.59  & 41.02 \\
		Karlsson		&51.25&57.40&\textbf{48.38}&60.97&60.97\\
		Heitz-Buschart		&\textbf{41.35}&41.77&42.12&72.66&46.71 \\
		\hline
	\end{tabular}
	\label{real_rmse}
\end{table}


We use Bootstrap method to quantify the uncertainty of predictions, where we draw Bootstrap samples from each study separately, and repeat the prediction procedures on the Bootstrapped studies. A total of 100 Bootstrap replicates are performed in our experiments. Figure \ref{3real} shows the Bootstrap log RMSE ratio boxplots. We observe consistent results with Table \ref{real_rmse}, where our proposed multi-source stacking DA algorithm has the smallest prediction RMSE over all methods when Qin or Karlsson study is used as the target domain, while when the Heitz-Buschart study is used as the target domain, the stacking algorithm has slightly larger median than the baseline merging method. 



\section{Discussion}
\label{discussion}

In this paper, we propose methods for multi-source DA in the regression setting. We first introduce a single-source DA algorithm for continuous outcomes that combines an extended BBSE algorithm and adversarial learning. We then  generalize our single-source DA to the multi-source setting through ensemble learning, where the target-adapted single source prediction models for each source-target domain pair are linearly combined to obtain the final predictions. We consider three strategies to select the combination weights: one based on multi-study stacking, one based on a source-target similarity measure and a combination of both.

We evaluate the performance of our proposed methods through extensive simulation studies and a real data application. Both experiments show that our multi-source stacking DA algorithm can substantially improve the prediction RMSE on the target domain as compared with either the merging approach or the normal stacking method where no DA is considered. The simulation results also demonstrate that the multi-source stacking DA algorithm is robust over varying degrees of study heterogeneity. 

Motivated by the real data application when Heintz-Buschart study is used as the target domain, our multi-source DA methods have larger prediction RMSE than the baseline merging approach, as a next step, we will explore conditions when DA might not be preferred over the simple merging approach, and therefore providing guidance on the choice of model to be trained for multi-source prediction.

\bibliographystyle{plain} 
\bibliography{ref}

\end{document}